\documentclass[3p,times,procedia]{elsarticle}
\usepackage{enumitem}
\usepackage{multirow}
\usepackage{mathbbol}
\usepackage{amsmath}
\usepackage{subcaption}
\usepackage{graphicx}

\usepackage{ecrc}
\usepackage[bookmarks=false]{hyperref}
    \hypersetup{colorlinks,
      linkcolor=blue,
      citecolor=blue,
      urlcolor=blue}
\volume{00}

\firstpage{1}

\journalname{Procedia Computer Science}

\runauth{Sirui Li et al.}

\jid{procs}

\usepackage{amssymb}
\usepackage{algorithmic}
\usepackage[linesnumbered,ruled,vlined]{algorithm2e}
\usepackage{subcaption}
\usepackage[figuresright]{rotating}
\begin{document}
\begin{frontmatter}



\dochead{26th International Conference on Knowledge-Based and Intelligent Information \& Engineering Systems (KES 2022)}%
%% e.g. \dochead{17th International Conference on Dynamical Processes in Excited States of Solids}

\title{Modelling Multi-relations for Convolutional-based Knowledge Graph Embedding}


\author[a]{Sirui Li} 
\author[a]{Kok Wai Wong}
\author[a,b]{Dengya Zhu}
\author[a]{Chun Che Fung}
%\author[a,b]{Third Author\corref{cor1}}

\address[a]{Discipline of Information Technology, Murdoch University, South St, Murdoch, Western Australia}
\address[b]{Discipline of Business Information Systems, School of Management and Marketing, Curtin University, Kent St, Bentley, Western Australia}

\begin{abstract}
Representation learning of knowledge graphs aims to embed entities and relations into low-dimensional vectors. Most existing works only consider the direct relations or paths between an entity pair. It is considered that such approaches disconnect the semantic connection of multi-relations between an entity pair, and we propose a convolutional and multi-relational representation learning model, ConvMR. The proposed ConvMR model addresses the multi-relation issue in two aspects: (1) Encoding the multi-relations between an entity pair into a unified vector that maintains the semantic connection. (2) Since not all relations are necessary while joining multi-relations, we propose an attention-based relation encoder to automatically assign weights to different relations based on semantic hierarchy. Experimental results on two popular datasets, FB15k-237 and WN18RR, achieved consistent improvements on the mean rank. We also found that ConvMR is efficient to deal with less frequent entities.
\end{abstract}

\cortext[cor1]{Corresponding authors: Sirui Li: sirui.li@murdoch.edu.au; Kok Wai Wong: K.Wong@murdoch.edu.au}
\end{frontmatter}

%\correspondingauthor[*]{Corresponding author. Tel.: +0-000-000-0000 ; fax: +0-000-000-0000.}
%\email{author@institute.xxx}

% \linenumbers


%\enlargethispage{-7mm}
\section{Introduction}
\label{intro}
A Knowledge Graph (KG) represents a graph-structured knowledge base to encode real-world entities and illustrate the relationship between them. There are two main branches: triple-based KGs and quadruple-based KGs. The former is typically represented as sets of Resource Description Framework (RDF) triples \textit{(s, r, o)}, where $s$ is the subject entity, $o$ is the object entity and $r$ is the relation, e.g. \textit{(Barack Obama, president of, USA)}. The latter is often called Temporal KGs (TKGs) represented as quadruples $(s, r, o, T)$, where $T$ denotes the timestamp, e.g. \textit{(Barack Obama, president of, USA, 2010)}. Although typical KGs contain millions of entities and billions of triples, they are far from complete~\cite{lin2015modeling}. %The methods to complete triple-based and quadruple-based KGs are slightly different because the latter additionally requires a timestamp analyser. 
In this paper, we focus on completing triple-based KGs %, such as Freebase~\cite{bollacker2008freebase} and WordNet~\cite{miller1995wordnet}, 
because the application of TKGs is limited to time-dependent domains, while triple-based KGs can be applied to more general domains~\cite{wu2020temp}.
%they are more popular than TKGs in many domains, e.g. health care~\cite{yu2017knowledge} and education~\cite{chi2018knowledge}. 
%Triple-based KGs are important resources in many Natural Language Processing (NLP) applications, e.g. question answering~\cite{li2021improving}. Although typical KGs contain millions of entities and billions of triples, they are far from complete~\cite{lin2015modeling}. 
%,

Many efforts have been devoted to predicting the missing links between subject-object pairs for triple-based KGs~\cite{bordes2011learning}. The traditional method using formal logic is neither traceable nor robust when dealing with long-range reasoning over a large scale KG~\cite{wang2014knowledge}. Recent studies~\cite{bordes2013translating, wang2014knowledge, dai2018novel, vu2019capsule} attempt to learn low-dimensional representations of entities and relations while preserving
certain properties of the original KG. The objective of these studies is to learn a score function that assigns higher plausibility scores to valid triples than invalid triples~\cite{dai2018novel}.

There are two main kinds of recent studies: triple-level learning and path-level learning. The triple level learning~\cite{bordes2013translating, wang2014knowledge, dai2018novel, vu2019capsule} only takes direct relations between entities into account. For example, TransE~\cite{bordes2013translating} assumes that the relation between two entities corresponds to a translation between the vector representations of the entities. %, that is, $\mathbf{v_s} + \mathbf{v_r} = \mathbf{v_o}$ when the triple \textit{(s, r, o)} holds. ConvKB~\cite{dai2018novel} keeps the translation of TransE but uses 1D convolutions to extract more feature interactions. 
The path-level learning~\cite{lin2015modeling, yang2017differentiable, guo2019learning} claims that there are also substantial multiple-hop relation paths between entities
indicating their semantic relationships. For instance, PTransE~\cite{lin2015modeling} is a path-based TransE, which not only learns vector representations from directly connected triples but also builds triples from KGs using entity pairs connected with relation paths.

Despite their success in predicting missing links, the triple/path-level learning methods disconnect the diverse aspects of multi-relations that are highly semantically related~\cite{zhang2017knowledge}. For an instance extracted from FB15k-237 as shown in Table~\ref{tab: hyper_ex}, there are a total of four relations between entity \textit{Prince Edward Island} and entity \textit{Canada}, which indicates the positional relationship at different hierarchies. In the triple-level learning, models split the example into four different triples and feed them to the score function $f$: $f$\textit{(Prince Edward Island, $r_a$, Canada)}, $f$\textit{(Prince Edward Island, $r_b$, Canada)}, $f$\textit{(Prince Edward Island, $r_c$, Canada)} and $f$\textit{(Prince Edward Island, $r_d$, Canada)}. In the path-level learning, models firstly mine paths between entity \textit{Prince Edward Island} and entity \textit{Canada}. Then models learn the score function based on the four triples and paths. Either triple-level learning or path-level learning treats this example as four different triples and therefore weakens the semantic connection between the four relations. Obviously, to reflect the relevant knowledge, it is more reasonable to model multi-relations between a pair of entities as an individual vector representation rather than divide them into four different triples~\citep{zhang2017knowledge}.
\begin{table}
\caption{An example of multi-relations in FB15k-237. $r_a$, $r_b$, $r_c$ and $r_d$ represent the four relations, respectively.}
\centering
\begin{tabular}{ccc}
\hline
\textbf{Subject} & \textbf{Relation}&\textbf{Object}\\
\hline
\multirow{ 4}{*}{Prince Edward Island} & $r_a$: /base/biblioness/bibs\_location/country & \multirow{ 4}{*}{Canada}\\
&$r_b$: /location/administrative\_division/country&\\
&$r_c$: /location/administrative\_division/first\_level\_division\_of&\\
&$r_d$: /base/areas/schema/administrative\_area/administrative\_parent&\\
\hline
\end{tabular}
\label{tab: hyper_ex}
\end{table}

Motivated by the observation, we propose a neural network-based schema, \textbf{ConvMR} (Convolutional and Multi-Relational model), to typically address the issue related to the multi-relations between an entity pair. In ConvMR, in addition to directly connected triples, we also build triples from KGs using multi-relations between entity pairs. %As shown in Figure~\ref{}, 
The triple/path-level learning only considers direct connections between an entity pair, building $(s, r_1, o), (s, r_2, o), ..., (s, r_N, o)$ and optimizing the objective. In contrast, ConvMR also builds a multi-relation triple $(s, r_1 \circ r_2 \circ ... r_N, o)$, where $\circ$ is an operation to join the multi-relations into a unified relation representation. Compared with triple-level learning, ConvMR enhances the semantic connection between multi-relations and entities. Compared with path-level learning, ConvMR relieves the burden to mine meaningful paths. There are two challenges to make ConvMR non-trivial. Firstly, the multi-relations should be properly encoded into a vector representation. The encoder is supposed to be easily implemented and maintain the semantic connection. Secondly, some relations could be less relevant while joining multi-relations because not every relation is necessary to the objective. To address the first challenge, we compared different encoders and chose the best one. To address the second challenge, we combined the chosen encoder with the attention mechanism. When dealing with variable sized inputs, the attention mechanism focuses on the most relevant parts of the input~\cite{DBLP:journals/tnn/GalassiLT21}.
%In summary, the contributions are:
%\begin{itemize}
%\item We addressed the semantic disconnection between the multi-relations, which is hardly considered by previous studies. We studied different encoders to embed the multi-relations into a unified relation representation regard of the
%\item We further the attention
%\end{itemize}
\section{Related Works}
\subsection{Triple-based Knowledge Graph Embedding}
Many Knowledge Graph Embedding models (KGEs) have been proposed to reason missing links between subject-object pairs. The key idea of recent KGEs is to learn low-dimensional vectors of entities and relations. 
\subsubsection{Triple-level learning}
Most of the currently available techniques~\cite{wang2014knowledge, bordes2013translating, trouillon2016complex, lin2015learning} use triples of a KG to perform the embedding task, enforcing vectors to be compatible with triples. These triple-level learning techniques can be roughly categorized into two groups: translational distance models and neural networks-based models. Translational distance models~\cite{bordes2013translating, wang2014knowledge, lin2015learning} view the relation as a translation operation between the subject and the object. For example, given a valid triple $(s, r, o)$, TransE~\cite{bordes2013translating} translates $r$ to a sum operation, $\mathbf{v_s} + \mathbf{v_r} = \mathbf{v_o}$, where $\mathbf{v_x}$ is the vector of item x. TransE is simple but it has flaws in dealing with complex relations, e.g. 1-to-N, N-to-1, and N-to-N relations~\cite{wang2017knowledge}. To better model complex relations, some models~\cite{wang2014knowledge, lin2015learning} allow the entity or the relation to have distinct representations. %For instance, TransH~\cite{wang2014knowledge} maps subject and object to the corresponding relation hyperplane. %TransR~\cite{lin2015learning} replaces hyperplanes with relation-based matrices that translate subject and object into relation space. 
However, Song et al.~\cite{song2020learning} argue that translational distance models suffer from slow convergence and poor local optimality. Moreover, these shallow models are limited to their expressiveness~\cite{xie2020reinceptione}.

Recent studies~\cite{dai2018novel, dettmers2018convolutional, balavzevic2019hypernetwork, jiang2019adaptive, vu2019capsule} have raised interest in applying deep neural networks to the triple-level learning. For example, ConvE~\cite{dettmers2018convolutional} is a Convolutional Neural Networks (CNNs)-based model, concatenating $\mathbf{v_s}$ and $\mathbf{v_r}$ into a matrix and the matrix is fed to the convolution layer. However, ConvE ignores the transitional characteristic between triples~\citep{dai2018novel}. To overcome this issue, Nguyen et al.~\cite{dai2018novel} propose ConvKB which concatenates $\mathbf{v_s}$, $\mathbf{v_r}$ and $\mathbf{v_o}$ together before feeding them to the convolution layer. Still, these neural networks-based models are triple-level learning methods so they often simplify the complex nature of the data in a KG~\cite{rosso2020beyond}.
\subsubsection{Path-level learning}
The triple-level learning methods only use direct relations between entities so they ignore rich information in relation paths~\cite{lin2015modeling}. On the other hand, the path-level learning methods~\cite{lin2015modeling, yang2017differentiable, guo2019learning} can explore relation paths between entities. For instance,  PTransE~\cite{lin2015modeling} improves TransE by incorporating relation inferences into KGEs. If there exists a path $(s_1, r_1, s_2, r_2, s_3)$ and a triple $(s_1, r_3, s_3)$, PTransE models a relation inference by learning $\mathbf{v_{r_1} \circ v_{r_2} = v_{r_3}}$ where $\circ$ is an operator, e.g. multiply. Yang et al.~\cite{yang2017differentiable} use first-order logical rules to generate paths. However, they ignore relational dependencies of entities and they do not fully exploit the potential of KG paths~\cite{guo2019learning}.

Some path-level methods~\cite{perozzi2014deepwalk, grover2016node2vec} use neural networks to generate paths. %, e.g. Path-Ranking Algorithm (PRA)~\cite{lao2010relational}. 
For example, DeepWalk~\cite{perozzi2014deepwalk} uses the uniform random walks to sample paths and employs Skip-Gram~\cite{mikolov2013linguistic} to encode these paths. On the other hand, neural networks are also used to capture the long-term relational dependencies of paths. Guo et al.~\cite{guo2019learning} propose Recurrent Skipping Networks (RSNs) to enhance semantic learning of relational paths. %Neelakantan et al.~\cite{neelakantan2015compositional} develop an Recurrent neural networks (RNN)-based model that applies compositionality recursively to compose the  implications of relational paths. 
While encoding relational paths improves model performance, complex algorithms have to be proposed to generate or prune meaningful paths~\cite{wang2017knowledge}.
\subsection{Quadruple-based Knowledge Graph Embedding}
Recent works of quadruple-based KGEs extend triple-based KGEs to the temporal domain. For example, some works~\cite{DBLP:conf/emnlp/DasguptaRT18, DBLP:conf/aaai/GoelKBP20, DBLP:conf/iclr/LacroixOU20} add a timestamp encoder and design time-sensitive quadruple decoding functions. However, entity-level temporal patterns are not explicitly captured by these methods~\cite{wu2020temp}. On the other hand, some works~\cite{manessi2020dynamic, DBLP:conf/aaai/ParejaDCMSKKSL20, DBLP:conf/nips/HajiramezanaliH19} use message passing networks to capture intra-graph neighbourhood information that is combined with the temporal recurrence or attention mechanisms. However, their focus is on continuous TKG~\cite{wu2020temp}.
\subsection{Attention Mechanism}
Attention mechanisms have become almost a \textit{de facto} standard in many Natural Language Processing (NLP) tasks~\citep{DBLP:journals/corr/BahdanauCB14}. Attention mechanisms are neural networks that could automatically weight the relevance of the input and take such weights into account in the downstream tasks. %For example, the machine translation model, e.g. RNNsearch~\cite{DBLP:conf/eacl/DyerHSYD17}, adds an attention function between the encoder and decoder. At each time step, the attention function assigns more important scores to the relevant context. With the help of the attention mechanism, the encoder-decoder can handle the long-range sentences and the vanishing/exploding gradient problem of RNNs. 
%Attention mechanisms are also popular to perform classification of the graph-structure data. 
%For instance, GAT~\cite{DBLP:conf/iclr/VelickovicCCRLB18} applied a masked attentional layer to indicate the importance of node j's feature to node i, where node j is a neighbour of node i. 
Recently, some KGEs~\cite{wang2019knowledge, li2021learning} incorporate the attention mechanism to enhance the effectiveness. HRAN~\cite{li2021learning} considers the importance of relational paths through the attention mechanism. Based on the
learned attention values for each relational path, HRAN aggregates the neighbour features in a hierarchical structure. Similarly, Wang et al.~\cite{wang2019knowledge} propose Graph Attenuated Attention networks (GAATs) which integrates an attenuated attention mechanism to assign different weights to different relational paths and acquire the information from the neighbours.
\section{Methodology}
A KG $\mathcal{G}=<E, R>$ is a collection of valid triples in the form of \textit{(subject, relation, object)} denoted by $(s, r, o)$, where $E$ is a set of entities and $R$ is a set of relations. %Suppose there is a total number of N relations connecting two entities: \textit{(s, $r_1$, o), (s, $r_2$, o), ..., (s, $r_N$, o)}. For each triple $(s, r_i, o)$, the .
To clarify, we denote $\mathbf{v_x}$ as the vector representation of item $x$ and denote the dimensionality of vector representation by $k$. There are three important modules in the proposed ConvMR: multi-relation triple generator, relation encoder and the objective learning.
\subsection{Multi-relation Triple Generator}
\begin{algorithm}
\SetAlgoLined
\KwIn{All triples of a KG $\mathcal{G}$.}
\KwOut{Original triples and multi-relation triples.}
pair2rel=dict()\\
multi-triples=set()\\
triples = all triples of $\mathcal{G}$\\
\For{$(s, r, o)$ in $triples$}{
\If{$(s, o)$ not in pair2rel}{
	 pair2rel[(s, o)] = set()
}
$pair2rel[(s, o)]$.add($r$)\\
}
filter = find all keys in pair2rel whose length of value is greater than 1\\
\For{$(s, o)$ in filter}{
$r_1, r_2, ..., r_N = pair2rel[(s, o)]$, where N is the number of relations between $(s, o)$\\
multi-triples.add($(s, r_1, r_2, ..., r_N, o)$)
}
return triples+multi-triples
 \caption{Algorithm for multi-relation triple generator.}
 \label{alg: generator}
\end{algorithm}
In learning, ConvMR concerns not only direct connections between an entity pair (i.e., original triples of a KG: $(s, r_1, o), (s, r_2, o), ..., (s, r_N, o)$) but also multi-relations between the pair (i.e., $(s, r_1, r_2, ..., r_N, o)$). We introduce the process of building the latter in Algorithm~\ref{alg: generator}. Generally, it uses a dictionary whose keys are $(s, o)$ pairs and values are the relation set.
\subsection{Relation Encoder}
Besides building multi-relation triples, ConvMR also needs a reliable relation encoder that can represent either one relation from an original triple or a set of relations from a multi-relation triple. In this paper, we consider four type of operations: set-transformer~\cite{lee2019set}, attention-based average (attn-average), Gated Recurrent Unit (GRU)~\cite{cho2014properties} and bi-directional GRU (biGRU). They are chosen due to the easy implementation and fast computation. To simplify the equation, we use $(s, H, o)$ to represent the input of each operation. The input can be either a triple $(s, r, o)$ or a multi-relation triple $(s, r_1, r_2, .., r_N, o)$.

\textbf{Set-transformer.} We assume that the sequence of the multi-relations does not matter in learning. Hence, we introduce the set-transformer~\cite{lee2019set} that is an attention-based neural network module. It is designed to model interactions among elements in the input set. It is formalized as:
\begin{equation}
Z = Encoder(H) = SAB(SAB(H))
\end{equation}
\begin{equation}
\mathbf{v_{r'}} = Decoder(Z) = rFF(SAB(PMA(Z)))
\end{equation}
\begin{equation}
SAB(H) = MAB(H, H)
\end{equation}
where $\mathbf{v_{r'}}$ is the representation of $H$, MAB is Multihead Attention Block, PMA is Pooling by Multihead
Attention, SAB is the Set Attention Block.

\textbf{Attn-average.} Attn-average stands for the attention-based average operation. Given the input $(s, H, o)$, attn-average firstly calculates the attention weight $a_i$ for each $r_i \in H$:
\begin{equation}
\mathbf{A} = softmax(tanh(\mathbf{W} \times \mathbf{H})) 
\end{equation}
The output $\mathbf{A} = [a_1, a_2, ..., a_N] \in \mathbb{R}^{1 \times N}$ is the attention weight matrix. Matrix $\mathbf{H} = [\mathbf{v_{r_1}, v_{r_2}, ..., v_{r_N}}] \in \mathbb{R}^{k \times N}$. $\mathbf{W}$ is a shared linear transformation. $tanh$ is the activation function. Then the attn-average represents $H$ by:
\begin{equation}
\mathbf{v_{r'}} = avg(\mathbf{H} \odot \mathbf{A}) =\frac{a_1\mathbf{v_{r_1}}+ a_2\mathbf{v_{r_2}} + ... + a_N\mathbf{v_{r_N}}}{N}
\end{equation}
where $\odot$ is the element-wise multiplication; $avg$ is the average operation.

\textbf{GRU.} Set-transformer or attn-average is based on the assumption that the sequence of the input is unnecessary. To investigate if the assumption is true, we employ GRU~\cite{cho2014properties} which is a gating mechanism in RNNs. GRU can capture the long-term dependency in sequences. It is formalized as:
\begin{equation}
\mathbf{v_{r'}} = GRU(\mathbf{H})
\end{equation}

\textbf{biGRU.} GRU only considers one direction of the input. A BiGRU is a sequence processing model that consists of two GRUs. One takes the input in a forward direction and the other in a backwards direction. It is formalized as:
\begin{equation}
\mathbf{v_{r'}} = biGRU(\mathbf{H})
\end{equation}
\subsection{Objective Formalization}
The objective of ConvMR (or other KGEs) is to learn a score function giving higher scores to valid triples than invalid triples. Given the subject vector $\mathbf{v_s} \in \mathbb{R}^{1 \times k}$, the object vector $\mathbf{v_o} \in \mathbb{R}^{1 \times k}$ and the relation/multi-relation vector $\mathbf{v_{r'}} \in \mathbb{R}^{1 \times k}$, we train a convolution layer to calculate the score for the matrix $\mathbf{T} = [\mathbf{v_s}, \mathbf{v_o}, \mathbf{v_{r'}}] \in \mathbb{R}^{k \times 3}$ that is the concatenation of $\mathbf{v_s}$, $\mathbf{v_o}$ and $\mathbf{v_{r'}}$. CNNs have shown its power to extract local features and generalize the transitional characters in previous KGEs~\cite{dai2018novel}. We use filters operated on the convolution layer. As a result, the score function $f$ is:
\begin{equation}
f(s, H, o) = concat(g([\mathbf{v_s}, \mathbf{v_o}, \mathbf{v_{r'}}] \ast \Omega ))
\end{equation}
where $concat$ is the concatenation operation, $\Omega$ is a set of filters, $\ast$ is the convolution operation; $g$ is the activation function ReLU.

We used the same loss function $\mathcal{L}$ as some previous KGEs, e.g. ConvKB~\cite{dai2018novel} and ComplEx~\cite{trouillon2016complex}. We used the AdaGrad to train ConvMR by minimizing the loss function with $L_2$ regularization on the weight vector $\mathbf{w}$ of the model.
The loss function $\mathcal{L}$ is:
\begin{equation}
\mathcal{L} = \sum_{(s, H, o) \in \mathcal{G}\cup\mathcal{G'}} log(1+
exp(l_{(s, H, o)} \cdot f(s, H, o)))+\frac{\lambda}{2}||\mathbf{w}||^2_2
\end{equation}
where $l_{(s, H, o)}=1$ for $(s, H, o) \in \mathcal{G}$; $l_{(s, H, o)}=-1$ for $(s, H,o) \in \mathcal{G'}$. $\mathcal{G'}$ is generated by corrupting valid triples in $\mathcal{G}$.
\section{Experiments}
\subsection{Datasets}
We evaluated the proposed ConvMR on two popular datasets: WN18RR~\cite{dettmers2018convolutional} and FB15k-237~\cite{toutanova2015observed}. The statistical information of WN18RR and FB15k-237 is provided in Table~\ref{tab: stat}.
\begin{table}
\caption{Statistics of the experimental datasets. \#E is the number of entities. \#R is the number of relations.}
\centering
\begin{tabular}{c|ccccc}
\hline
\textbf{Dataset} & \textbf{\#E} & \textbf{\#R} & \multicolumn{3}{c}{\textbf{\#Triples in train/valid/test}}\\
\hline
WN18RR&40,943&11&86,835&3,034&3,134\\
FB15k-237&14,541&237&272,115&17,535&20,466\\
\hline
\end{tabular}
\label{tab: stat}
\end{table}
\subsection{Baselines}
We included eight baseline techniques and most of them are CNNs-based KGEs because the proposed ConvMR employed CNNs to train the score function.
\begin{itemize}
\item TransE~\citep{bordes2013translating} is a popular transition-based method. TransE assumes the subject vector representation plus the relation vector representation should be close to the object vector representation.
\item ConvE~\citep{dettmers2018convolutional} is the first KGE using CNNs. ConvE finds the convolutional features in the concatenation of the subject vector representation and the relation vector.
\item ConvKB~\citep{dai2018novel} improves ConvE by concatenating subject, relation and object before feeding them into the convolution layer.
\item HypER~\citep{balavzevic2019hypernetwork} is a variation of ConvE. HypER introduces hypernetworks to generate relation-specific filters for the convolution layer.
\item ConvR~\citep{jiang2019adaptive} is a modification of ConvKB. Instead of using global filters like ConvKB, ConvR performs an adaptive convolution with relation-specific filters.
\item CapsE~\citep{vu2019capsule} improves ConvKB by adding a ``deep" architecture, capsule (each capsule is a group of neurons), to capture the intrinsic spatial relationship between triples.
\item RSN~\citep{guo2019learning} concentrates on learning vector representations from relational paths by employing Recurrent Neural Networks (RNN).
\item CompGCN~\citep{DBLP:conf/iclr/VashishthSNT20} addresses the issue of GCNs that they only learn entity embeddings. CompGCN jointly learns node representations and relation representations in the graph to solve the KG sparsity.
\end{itemize}
\subsection{Training Protocol}

In training, we randomly replaced $s$ or $o$ to generate invalid instances. We employed the same initialization of entity and relation vector representations as CapsE~\citep{vu2019capsule}. We tuned the hyper-parameters with the grid search. Finally, we trained ConvMR for 20 epochs and set $k$=100, the number of batch was 800, learning rate of the AdaGrad optimizer started from 0.01 and was reduced by 0.1/5 epochs. 
\subsection{Evaluation Metrics}

%The input of ConvHR is triples $(s, r, o)$ in evaluation for the fair comparison with baselines. 
Previous works perform the link prediction task to evaluate KGEs. Given a relation and an entity, the purpose of the link prediction task is to predict a missing entity, i.e., referring to $s$ given $(r, o)$ or inferring to $o$ given $(s, r)$. To generate the corrupted triples in the link prediction task, we followed the ranking process proposed in~\citep{bordes2013translating}: for each triple $(s, r, o)$, we replaced either $s$ or $o$ with every entity $e \in E$. We use the Filtered setting protocol~\citep{bordes2013translating}, i.e., not taking any corrupted triples that appear in the KG into accounts. We employed evaluation metrics: Mean Rank (MR) (the average of predicted ranks) and Hits@10 (the proportion of correct entities ranked in top 10). A lower MR and a higher Hits@10 value is the mark of a better model. Note that the input is triples $(s, r, o)$ in test for the fair comparison with baselines.
\subsection{Results} 
We firstly compared the four types of operations in the relation encoder and chose the best one. Then, we designed three experiments to evaluate the power of learning the joint of multi-relations. Finally, we compared the proposed ConvMR with baselines.
\subsubsection{Comparison of Different Relation Encoders}
For ConvMR, we considered four operations for the relation encoder: set-transformer, attn-average, GRU and biGRU. From Table~\ref{tab: operator} we observed that: (1) Overall, attn-average outperformed other operations on all metrics across both datasets. While the set-transformer had the worst MR and lowest Hits@10 on both datasets. There are two reasons leading to the major gap between the two operations. Firstly, the set-transformer trained more parameters than the attn-average, which increased the difficulty of training and hence downgraded the overall performance. Secondly, the set-transformer aggregated features of multi-relations into a dense vector by using complex components, such as the multi-head attention block. It might constrain the ability of CNN to extract local features or to generalize the transitional character from triples. (2) Attn-average did better than GRU and biGRU, especially on FB15k-237 where attn-average gained significant improvements of 366-155=211 in MR and 52.5-30.3=22.2 in Hits@10. It represented that finding the long dependency of multi-relations did not contribute to the relation encoder. (3) To investigate whether the sequence or direction of the multi-relations is necessary, we made a comparison between GRU and biGRU. Obviously, biGRU did not dramatically outperform GRU. They achieved similar MR and Hits@10 on both datasets. It showed that the sequence or direction of multi-relations is unnecessary. Based on the above analysis, we chose the attn-average as our relation encoder in the following experiments.
\begin{table*}[h]
\caption{Comparison results of four operations in the relation encoder.}
  \centering
  \begin{tabular}{c|cc|cc}
\hline
\multirow{2}{*}{Operation} & \multicolumn{2}{c|}{WN18RR} & %
    \multicolumn{2}{c}{FB15k-237} \\
\cline{2-5}
 & MR & Hits@10 & MR & Hits@10 \\
\hline
set-transformer &915 &45.4 &500 & 23.1\\
attn-average &\textbf{692} & \textbf{53.3}&\textbf{155} & \textbf{52.5} \\
GRU &720 & 50.1& 365 & 30.3\\
biGRU &718 & 50.1&366 &31.5 \\
\hline
\end{tabular}
  \label{tab: operator}
\end{table*}
\subsubsection{Effect of the Attention Mechanism}\label{sec: atten_eval}
In the following experiments, the relation encoder referred to attn-average. In this section, we studied how the attention mechanism affects the performance of the relation encoder by conducting an ablation study. We removed the attention mechanism of the relation encoder. In this case, multi-relations were only encoded by the average operation. On WN18RR, removing the attention mechanism decreased MR by 729-692=37, Hits@10 by 53.3-50.9=2.4\% . On FB15k-237, the removal leaded to a 211-155=56 decrease in MR and 52.5-46.4=6.1\% decrease in Hits@10. Therefore, the attention mechanism is important in the relation encoder.

To further demonstrate how the attention mechanism enhanced the performance of the relation encoder, we selected the highest frequency multi-relation from each dataset. The multi-relations with the highest frequency are (``derivationally\_related\_form", ``synset\_domain\_topic\_of") in WN18RR and (``award\_winner", ``award\_nominee") in FB15k-237. We fed them into our trained attention mechanism. In WN18RR, the attention mechanism assigned 0.58 weight to ``derivationally\_related\_form" and assigned 0.42 weight to ``synset\_domain\_topic\_of" while encoding the multi-relations. In FB15k-237, the attention mechanism assigned 0.49 weight to ``award\_winner" and assigned 0.51 weight to ``award\_nominee".

A study~\cite{allen2019understanding} analysed the relations of popular KGs and classified them into three types based on the relatedness, suject specifics and object specifics: highly related (R), (generalised) specialisation (S) and (generalised) context-shift (C). Type C is a generalised case of type R and type S. That is, relations in type C have higher semantic relatedness, more subject and object specifics than relations in type R and type S. In WN18RR, the relation ``derivationally\_related\_form" belongs to type R and the relation ``synset\_domain\_topic\_of" belongs to type C. In FB15k-237, the vast majority of relations belong to type C. Based on this observation, we can conclude that the weights reflect the hierarchies of relations. Relation ``derivationally\_related\_form" is the parent hierarchy of relation ``synset\_domain\_topic\_of". Relation ``award\_winner" and relation ``award\_nominee" have the same or similar hierarchy. This finding supported that the relation encoder can maintain the semantic/hierarchical connection between relations by automatically assigning different weights.
\subsubsection{Removal of Multi-relation Triple Generator}\label{sec: exp_hyper}
In this section, we focused on the research question ``Do multi-relations contribute to the proposed model?". To answer this question, we trained ConvMR without the multi-relation triple generator. That is, the input of ConvMR was original triples from the KG. As a result, ConvMR without multi-relations got 735-692=43 MR and 53.3-51.0=2.3\% Hits@10 decreases in WN18RR; 200-155=45 MR and 52.5-50.4=2.1\% Hits@10 decreases in FB15k-237. The decreases indicated that learning with multi-relations captures more semantically rich neighbourhood information and hence produces more reasonable vector representations.
\begin{table}[h]
\caption{Comparison results of four categories. (with/without multi-relations)}\label{tab: gen}
\centering
 \begin{tabular}{c|cc|cc}
\hline
\multirow{2}{*}{Category} & \multicolumn{2}{c|}{WN18RR} & %
    \multicolumn{2}{c}{FB15k-237} \\
\cline{2-5}
 & MR & Hits@10 & MR & Hits@10 \\
\hline
1-1 &4/5 &96.4/96.4&292/300&54.9/53.4\\
1-M &839/883 &31.8/30.8&355/350& 36.2/35.7\\
M-1 &1157/1229 &30.6/27.5&208/230& 57.3/54.8\\
M-M &32/35&96.3/94.5&118/130& 51.2/48.8\\
\hline
\end{tabular}
\end{table}

Following~\citep{bordes2013translating}, we categorized relations into four categories based on the average number $n_s$ of subjects per objects and the average number $n_o$  of objects per subjects: one-to-one (1-1), one-to-many (1-M), many-to-one (M-1) and many-to-many (M-M). We compared the performance of ConvMR with and without the generator in each category. The results are shown in Table~\ref{tab: gen}. Overall, there was no significant difference in 1-1 and 1-M on MR and Hits@10 across both datasets. However, ConvMR with multi-relations gained obvious improvement on M-1 and M-M on both datasets. Entities in M-1 and M-M triples often appear less frequently than entities in 1-1 and 1-M triples~\citep{vu2019capsule}. In summary, training ConvMR with multi-relations can improve the representation learning, especially for rare entities.
\subsubsection{Comparison with Baselines}
\begin{table*}
\caption{Results on WN18RR and FB15k-237 test sets. Results of TransE, ConvE, ConvKB and CapsE are from~\citet{vu2019capsule} where TransE, ConvKB and CapsE used the same vector representation intiailization as ConvMR; CompGCN and HypER are from the original papers; RSN and ConvR are from \citet{rossi2021knowledge}}
  \centering
  \begin{tabular}{c|cc|cc}
\hline
\multirow{2}{*}{Method} & \multicolumn{2}{c|}{WN18RR} & %
    \multicolumn{2}{c}{FB15k-237} \\
\cline{2-5}
 & MR & Hits@10 & MR & Hits@10 \\
\hline
ConvR & 5646&52.7&251&52.6\\
RSN &4210 &48.3 &248 &44.4 \\
CompGCN & 3533&54.6 &197 & 53.5\\
HypER & 5798 &52.2 &250 &52.0 \\
ConvE&4187 &51.5 &244 &50.1 \\
\hline
TransE & 743 & \textbf{56.0} &347 &46.5 \\
ConvKB & 763 &\textbf{56.0}&254 &53.2 \\
CapsE &719 &\textbf{56.0}&303 &\textbf{59.3} \\
\hline
%ConvHR-gru & &  &&\\
ConvMR &\textbf{692} & 53.3&\textbf{155} &52.5 \\
\hline
\end{tabular}
  \label{tab: baseline}
\end{table*}
Table~\ref{tab: baseline} compared the experimental results of ConvMR with baselines, using the same evaluation protocol. From the results, we can see that: (1) ConvMR obtained the best MR and proper Hits@10 on both datasets. On WN18RR, ConvMR achieved 719-692=27 MR increase and outperformed four models on Hits@10. On FB15k-237, ConMR beat four models on Hits@10 and achieved 197-155=42 MR increase compared with all baselines but 250-155=95 increase compared with CNN-based KGEs. It is mainly because previous KGEs processed triples independently, disconnecting the semantic connection between multi-relations. Additionally, there are more M-1 triples (20.5\%) and M-M triples (72.3\%) in FB15k-237 than those (47.4\%, 36.1\%) in WN18RR, so the MR of FB15k-237 increased dramatically. In Section~\ref{sec: exp_hyper}, we have shown that training ConvMR with multi-relations performed better on M-1 and M-M triples even though entities in M-1 and M-M are rare. Moreover, our attention mechanism is able to specify importance weight to relations based on semantic hierarchy, which has been shown in Section~\ref{sec: atten_eval}. 
\section{Conclusion}
In this paper, we proposed a CNN-based KGE, ConvMR. Previous KGEs were triple/path-level learning methods, disconnecting the semantic connection of multi-relations between subject-object pairs. The proposed ConvMR addressed this issue by building multi-relation triples and learning the weighted joint of them. The experiments have presented two important findings: (1) Learning representations with multi-relation triples can effectively mine features of less frequent entities. (2) The weighted joint of multi-relations can automatically assign different weights to relations based on the hierarchy, which maintains the semantic connection between relations. Experimental results demonstrated that ConvMR outperforms eight baseline techniques on MR across two popular datasets: WN18RR and FB15k-237. Even though ConMR did not achieve the state-of-the-art performance on Hits@10, our work identified a new perspective of KGE. In the future, we will explore the generalization of ConvMR to quadruple-based KGs and apply ConvMR to other NLP applications.
\bibliographystyle{unsrtnat}

\end{document}